\pdfoutput=1
\documentclass[a4paper, 10pt, conference]{ieeeconf}      

\usepackage{FG2019}
\usepackage{graphicx}                   

\FGfinalcopy 

\IEEEoverridecommandlockouts                              
\title{\LARGE \bf
Micro-expression detection in long videos using optical flow and recurrent neural networks
}

\author{\parbox{16cm}{\centering
    {\large Michiel Verburg and Vlado Menkovski}\\
    {\normalsize
    Department of Mathematics and Computer Science, Eindhoven University of Technology, Eindhoven, The Netherlands}}\\
}

\begin{document}

\ifFGfinal
\thispagestyle{empty}
\pagestyle{empty}
\else
\author{Anonymous FG 2019 submission\\ Paper ID \FGPaperID \\}
\pagestyle{plain}
\fi
\maketitle

\begin{abstract}


Facial micro-expressions are subtle and involuntary expressions that can reveal concealed emotions. Micro-expressions are an invaluable source of information in application domains such as lie detection, mental health, sentiment analysis and more. 
One of the biggest challenges in this field of research is the small amount of available spontaneous micro-expression data. However, spontaneous data collection is burdened by time-consuming and expensive annotation. Hence, methods are needed which can reduce the amount of data that annotators have to review.
This paper presents a novel micro-expression spotting method using a recurrent neural network (RNN) on optical flow features. We extract Histogram of Oriented Optical Flow (HOOF) features to encode the temporal changes in selected face regions. Finally, the RNN spots short intervals which are likely to contain occurrences of relevant facial micro-movements.
The proposed method is evaluated on the SAMM database. Any chance of subject bias is eliminated by training the RNN using Leave-One-Subject-Out cross-validation. Comparing the spotted intervals with the labeled data shows that the method produced 1569 false positives while obtaining a recall of 0.4654.
The initial results show that the proposed method would reduce the video length by a factor of 3.5, while still retaining almost half of the relevant micro-movements. Lastly, as the model gets more data, it becomes better at detecting intervals, which makes the proposed method suitable for supporting the annotation process.

\end{abstract}

\section{INTRODUCTION}

Facial expressions constitute an important part of our daily social interactions, because they give insight into how a person feels. Facial micro-expressions are subtle and involuntary occurrences of expressions, lasting less than 1/2 a second \cite{Yan2013a}. Micro-expressions occur when emotions are being concealed \cite{ekman1969}, either deliberately or unintentionally, effectively leaking information about the concealed emotion. 

Recognizing micro-expressions can prove helpful in various application domains, from lie detection to sentiment analysis to assisting psychologists in mental health. Micro-expressions are an invaluable source of information because they reveal things we would not have easily known otherwise. In lie detection applications, missing a single clue could prove detrimental to understanding the true motive behind someone's actions. Moreover, in the domain of mental health, accurately recognizing micro-expressions can provide important information, for example, to decide whether a person is a danger to self and others. Currently, many micro-expressions are easily overlooked due to the low intensity and short duration of the expression, which is the reason this task is challenging for both humans and machines. A study by Frank et al. \cite{frank2009see} reported the accuracy for correctly spotting and classifying a micro-expression to be about 47\% for trained experts. Therefore, domain experts certainly could benefit from automatic micro-expression analysis to support them in their tasks.

The Facial Action Coding System (FACS) was developed to provide a way to label facial movements objectively \cite{Ekman1976}. The FACS defines a code, called an Action Unit (AU), for every type of facial movement. Using this system, dataset annotators can then label micro-expressions with the corresponding code based on the observed facial movement, rather than labeling it with a subjective classification of emotion. This allows micro-expressions to be defined more accurately, and consequently to be recognized more accurately by automatic systems.

There is a scarce number of spontaneous micro-expression databases. The currently available spontaneous databases are: SMIC \cite{Li2013}, CASME \cite{Yan2013}, CASME II \cite{Yan2014}, CAS(ME)$^2$ \cite{Qu2017}, and SAMM \cite{Davison2018}. However, only the SAMM dataset and the range of CASME databases are labeled with AUs rather than just the emotion classes.

In recent years, micro-expression research has been growing in popularity. However, most of the works have focused on recognition of micro-expressions, rather than the detection. The lack of annotated spontaneous micro-expression data limits the progress that can be made in this field. To overcome this issue, large quantities of footage would be required in which people might exhibit micro-expressions. After having obtained such footage, it would need to be annotated with the onset, apex (peak) and offset frames of the occurring micro-expressions, in addition to the AUs corresponding to each facial movement. This poses another challenge because it is quite costly to annotate large amounts of spontaneous micro-expression data. Currently, this annotation is performed by experts manually reviewing the footage to spot and classify micro-expressions. However, the sparseness of the spontaneous data requires the human annotators to look through long segments of footage which may contain only a few micro-expressions.
To diminish this problem, micro-expression spotting methods could be employed to remove segments that do not contain relevant facial micro-movements. To create such methods, we need datasets which contain labeled long video sequences.

\begin{figure}[thpb]
  \centering
  \includegraphics[width=0.45\textwidth]{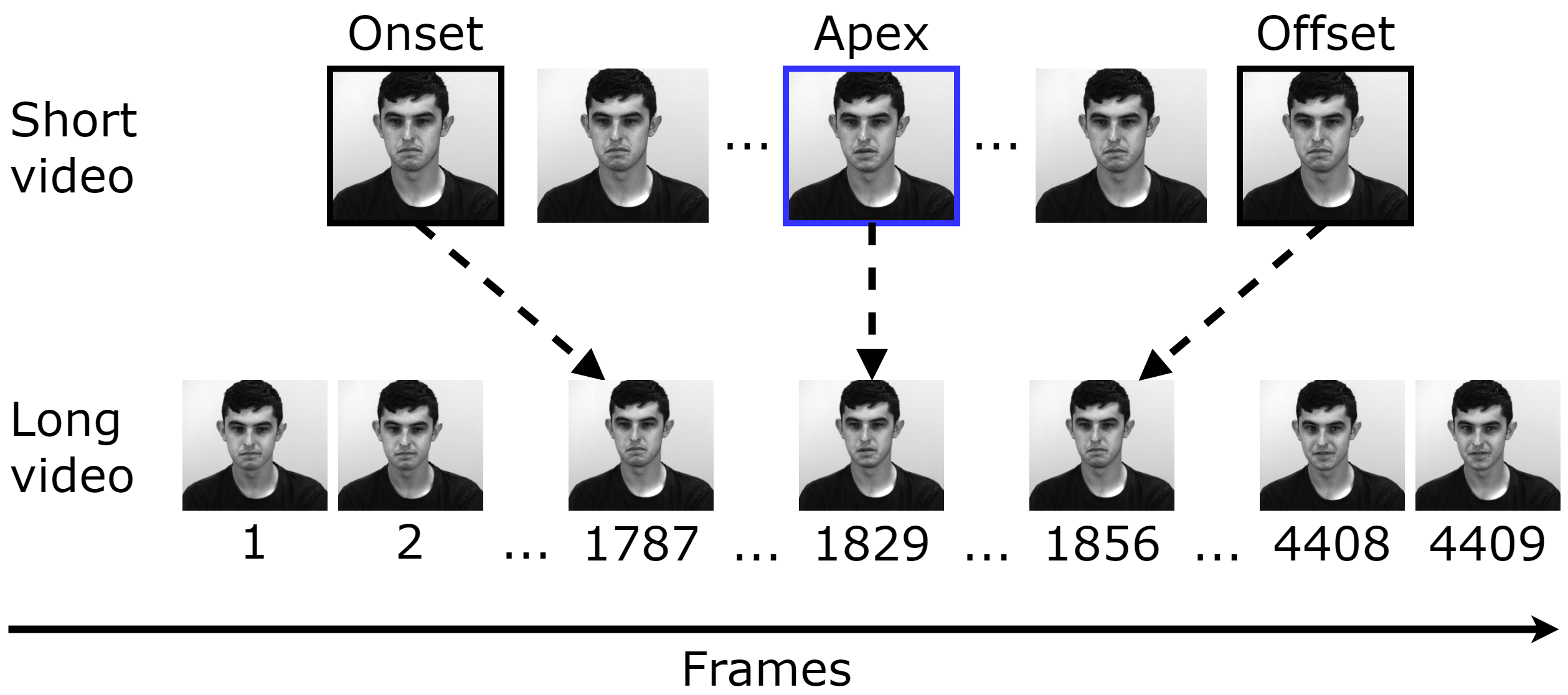}    
  \caption{Example of a short video with labeled onset, apex and offset frame, and an example of a long video that encompasses the short video. The sequence is extracted from the SAMM dataset, which was recorded at 200 fps \cite{Davison2018}}
  \label{fig:longvideo}
\end{figure}

Presently, the CAS(ME)$^2$ and SAMM databases are the only datasets that have published such long video sequences. The other spontaneous micro-expression databases provide (preprocessed) short videos which are limited to one micro-expression. In Fig. \ref{fig:longvideo} we show an example of such a ``short video'', and a ``long video''.

At the time of writing, only a limited number of published works have explored spotting in long video sequences \cite{Li2018a}, \cite{Wang2017}, \cite{Gupta2018}, \cite{Zhang2018}. However, there have been other works that explored micro-expression spotting, but since they did not have suitable long video data available to them yet, it is difficult to tell how those methods would perform on the long video sequences provided in CAS(ME)$^2$ and SAMM. Hence, there is still a large gap in research on micro-expression spotting in long videos.

In this work, we propose a method that will be able to spot micro-expression intervals in long video sequences by using a recurrent neural network (RNN) on optical flow features. We extract the Histogram of Oriented Optical Flow feature to encode the temporal change in selected face regions, which is then passed to the RNN for the detection task. We choose to use a recurrent neural network which is composed of long short-term memory (LSTM) units.
The method spots short intervals of 0.5s that are likely to contain micro-expressions. These preprocessed short intervals can be presented to an annotator for verification, which reduces the amount of footage that needs to be reviewed. Additionally, the annotator's decision on whether an interval contains a micro-expression or not can be fed back into the network, automatically improving the model's performance. The main contribution of this work is the proposed method's potential to make annotation of spontaneous micro-expression data less time-consuming, which will lower the threshold for data collection at larger scales.

We evaluate the method on the SAMM dataset, which has 159 labeled micro-movements, and consists of both micro-expressions and other relevant micro-movements. To acquire unbiased results over the entire dataset, the network is trained using Leave-One-Subject-Out cross-validation.
The proposed method produced 1569 false positives --- which amounts to a bit more than a quarter of the footage of the entire dataset --- while retaining almost half of the occurring micro-movements with the method's achieved recall of 0.4654. The initial results support the idea that a neural network could be utilized as a tool to make the annotation process easier.

The remainder of the paper is structured as follows: Section \ref{sec:related} discusses the related works; Section \ref{sec:method} introduces the proposed method; Section \ref{sec:results} presents and summarizes the results of the proposed method; finally, Section \ref{sec:conclusion} concludes this paper and discusses possibilities for future work.

\section{RELATED WORK}\label{sec:related}

\subsection{Region extraction}
Region extraction is a typical preprocessing step in micro-expression analysis frameworks. The motivation for extracting regions is to eliminate data that does not add enough value to the task of spotting micro-expressions.

Liong et al. \cite{Liong2015a} proposed the extraction of Regions of Interest (ROI) to spot the apex frame in a micro-expression sequence. The regions they proposed to extract are the ``left eye+eyebrow'', ``right eye+eyebrow'' and ``mouth'' regions. However, in a later work, Liong et al. \cite{Liong2017} proposed to mask the eye region, because eye blinks cause a lot of false positives.

Davison et al. \cite{Davison2018a} employed FACS-based regions, which involves selecting regions of the face that contain movements from particular Action Units (AUs). This allows for direct links between facial movements and specific AUs, which is subsequently used for detecting occurrences of micro-expressions.

\subsection{Micro-expression spotting}
Numerous works exist on automatic micro-expression analysis, however, only a limited amount has focused on micro-expression spotting. We aim to outline some of the important works on micro-expression spotting in this section.

Moilanen et al. \cite{Moilanen2014} introduced the appearance-based feature difference analysis method for micro-expression spotting. This method uses a sliding window of size $N$, where $N$ is equal to the average length of a micro-expression. The features of the center frame are compared with the average feature frame of the sliding window, which is the average between the features of the first and last frame of the window. The idea, then, is that if the window overlaps with a micro-expression (especially if the center frame is the peak of the micro-expression), the difference between the average feature frame and the feature of the center frame will be larger than when the window contains either no movement, or a macro-movement. The idea behind it is that a micro-expression will go from a neutral state to a peak, and then back to neutral again, all within the window of $N$ frames, thus causing a larger difference. The difference is generally calculated by using the Chi-Squared (${\chi}^2$) distance method on a pair of histogram-based features. Examples of features that have been used with this method are Local Binary Pattern (LBP) \cite{Li2017}, Histogram of Oriented Optical Flow (HOOF) \cite{Li2017}, and 3D Histogram of Oriented Gradients (3D HOG) \cite{Davison2018a}.
The result of this feature difference analysis is a feature difference vector over the frames of a given sequence. The local maxima of the feature vectors are typically peaks of some kind of subtle movement, which could be a micro-expression. The idea is then to take a certain threshold, such that if a peak is above the threshold, it likely corresponds to a micro-expression.

\section{PROPOSED METHOD}\label{sec:method}
The proposed method contains three main parts: preprocessing, feature extraction and micro-expression spotting. Preprocessing consists of cropping and aligning the face, followed by extracting regions of interest. Feature extraction consists of applying optical flow methods on the frames and using the resulting flow vectors to compute the HOOF feature.
Lastly, to spot micro-expression intervals, we feed the extracted features of a video sequence into an RNN composed of LSTM units, which then classifies whether a given sequence contains a relevant micro-movement, followed by a post-processing step which suppresses neighboring detections.

\subsection{Preprocessing}
\label{sec:preprocessing}
\begin{figure}[thpb]
  \centering
  \includegraphics[width=0.45\textwidth]{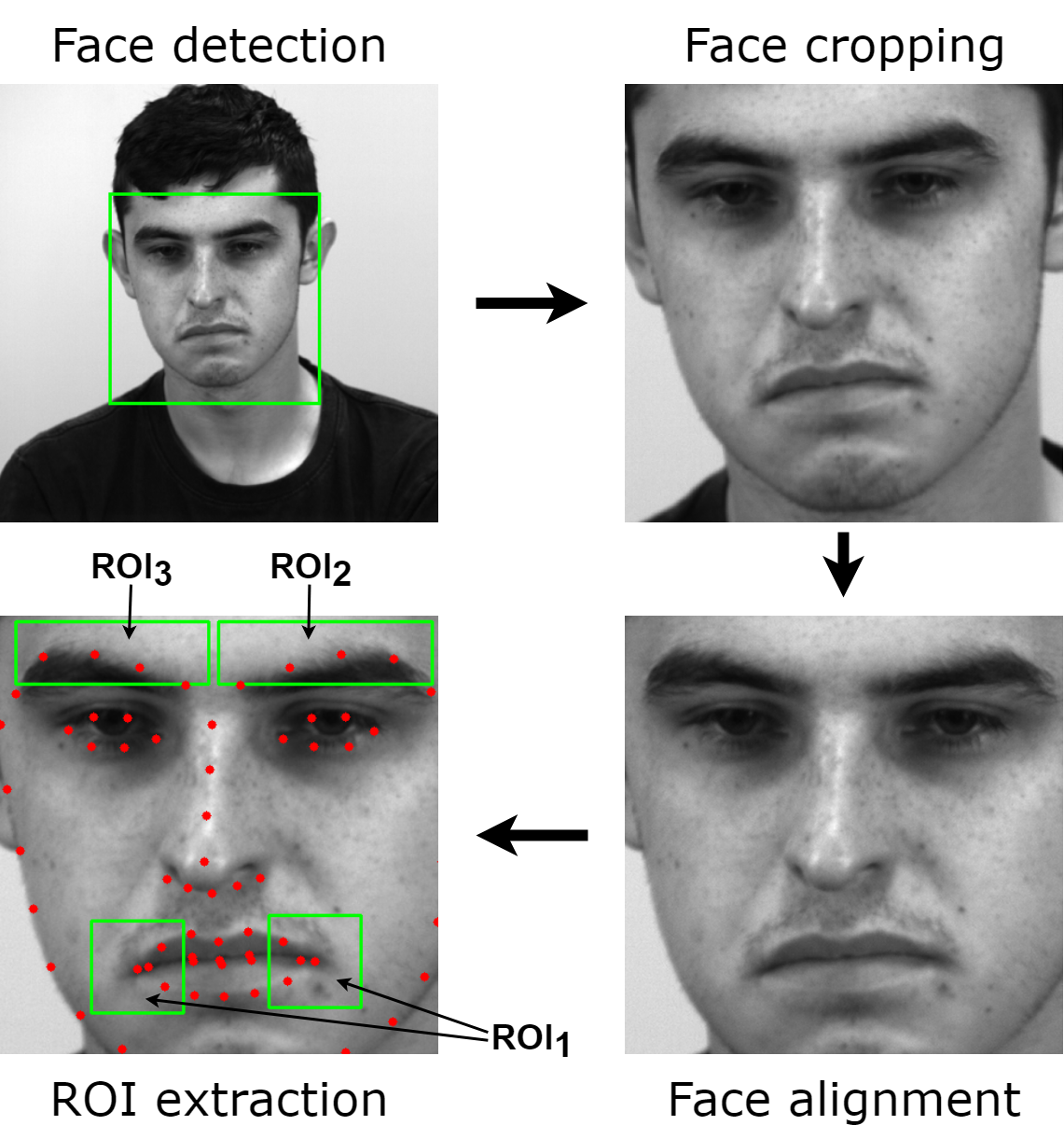}    
  \caption{The main steps in preprocessing raw face images.}
  \label{fig:preprocessing}
\end{figure}
We first detect the face using the default face detector in Dlib \cite{king2009dlib}, resulting in a bounding box that surrounds the face, which we then use to crop the face, as shown in the top half in Fig. \ref{fig:preprocessing}.

Secondly, we use a convenience function from the imutils package \cite{imutils}, which first localizes the facial landmarks on the cropped image using a pre-trained facial landmark detector within the Dlib library. The convenience function then calculates the center points of both eyes using the detected landmarks. Subsequently, the face is aligned such that the eyes are on a horizontal line, as shown in the bottom right of Fig. \ref{fig:preprocessing}. 

The last preprocessing step involves extracting regions of interest (ROI). To extract the ROIs, we have to refit the facial landmarks on the aligned image. We then use these facial landmarks to extract ROIs by selecting a bounding box around the corresponding landmarks, as shown in the bottom left of Fig. \ref{fig:preprocessing}. We have chosen to exclude the eye regions because eye blinking generates too much noise \cite{Shreve2009} \cite{Liong2017}, and we have chosen to exclude the nose region because it is rigid and provides little information \cite{Shreve2011}. Additionally, we exclude the middle region of the mouth, because experiments showed that this middle region only generated noise. Hence, we centered the bounding boxes around the corners of the mouth.

Since we are dealing with long videos, we need to somehow ensure that the face maintains the same orientation throughout the video. One way of doing this would be to align each frame independently, ensuring the eyes are on a horizontal line. However, this method is counterproductive since small errors in landmark localization cause unnecessary transformations of the entire head. The second method is to employ a temporal sliding window, as used by Li et al. \cite{Li2018a}. The first frame of the sliding window is used to localize the landmarks and to align the face, and consequently, the same transformation is applied to the remaining frames in the sliding window. Additionally, we extract the regions of interest based on the first frame of the window. By taking the size of the sliding window small enough, there will not be any significant changes in head orientation between the start and the end. In our case, we take the size of the sliding window $W$ to be 0.5s, which means that $|W|=100$ for 200 fps data. We choose this size for the sliding window because it is the upper bound on the duration of a micro-expression.

\begin{figure}[thpb]
  \centering
  \includegraphics[width=0.45\textwidth]{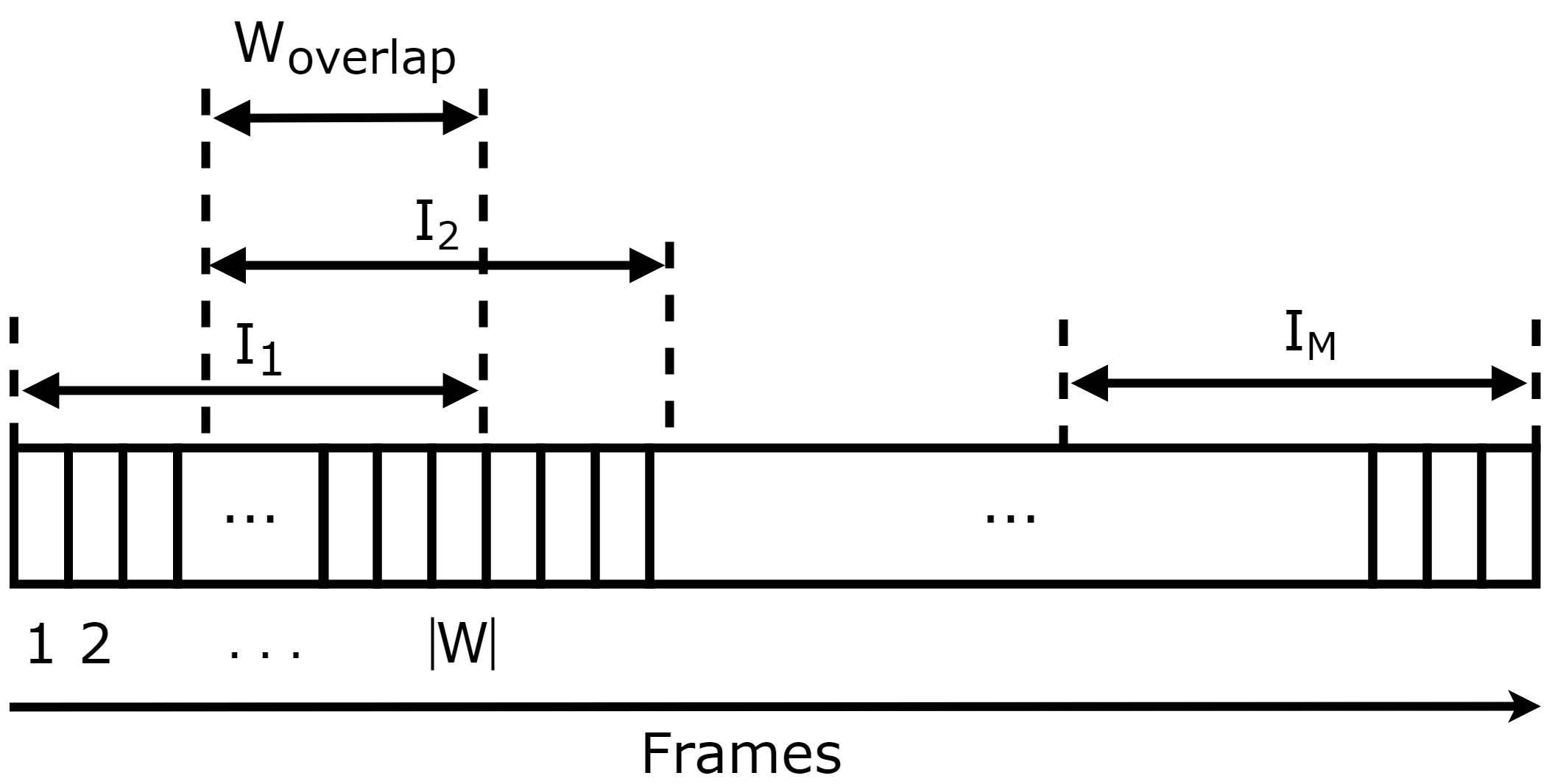}    
  \caption{Sliding window method used to generate an ensemble of $M$ intervals with length $|W|$ from a long video sequence. The intervals can then be preprocessed independently before being passed to the next stage.}
  \label{fig:window}
\end{figure}

Using the sliding window, an ensemble of intervals can be generated. However, this creates the possibility that a micro-expression would span two intervals, and in the worst case, both intervals would only contain 50\% of the micro-expression. Using only half of the micro-expression would make reliable detection difficult for the neural network. Therefore, we use an overlap $W_{overlap}$ to generate an ensemble of intervals $I_1,I_2,\ldots,I_M$ as shown in Fig. \ref{fig:window}. By taking the size of $W_{overlap}$ to be 0.3s, the sliding window method is guaranteed to generate intervals in a way such that for any micro-expression interval $S_{ME}$, there exists an interval $I_j$, $1 \leq j \leq M$, such that:
\begin{equation}
\label{eq:true}
\frac{I_j \cap S_{ME}}{S_{ME}} \geq 0.8
\end{equation}
Where we pick 0.8 because we believe that the network should be able to detect micro-expressions when given a sequence containing only 80\% of the interval $S_{ME}$.

\subsection{Feature extraction}
\label{sec:feature_extraction}
In the feature extraction stage, we make use of the Histogram of Oriented Optical Flow (HOOF) feature, which is an optical flow-based motion feature that extracts meaningful information for the spotting of micro-expressions. This stage is required because an LSTM learns dependencies in time, so it needs some kind of representation when we are using image data. In many cases, a Convolutional Neural Network (CNN) would be used to extract such representations automatically as well. However, due to the small amount of training samples, we explore the use of a handcrafted feature, because it gives a higher signal-to-noise ratio in comparison to using raw pixel data.

\subsubsection{Optical flow}
Optical flow-based features have been gaining more popularity in recent research \cite{Li2017}, which is partly because of its effectiveness in measuring spatio-temporal changes in intensity \cite{Oh2018}. The optical flow method calculates the difference between two frames by encoding the velocity and direction of each pixel's movement between the frames.

Optical flow has three main assumptions \cite{Liong2015}: it assumes the brightness is constant; it assumes that pixels in a small block originate from the same surface and have similar velocity; it assumes that objects change gradually and not drastically over time. Overall, these assumptions are typically met, because the face changes gradually over time; pixels in a small block typically come from the same surface; brightness, especially in the datasets, is kept quite constant. In real life, brightness might not always be constant, but most likely it will be constant enough for the duration of a (micro-)expression.

\subsubsection{Histogram of Oriented Optical Flow}
To extract features from each of the short intervals $I_1,I_2,\ldots, I_M$, generated as described in Section \ref{sec:preprocessing}, we compute the HOOF values between frames in the sequence. However, when we have high fps data, the optical flow between subsequent frames is insignificant. Therefore, we essentially downsample the data by a rate of $R$, where we take $R$ to be 1/50s, equivalent to 4 frames for 200 fps data. 
For a short video $V_{short}=f_1,f_2,\ldots,f_{W}$, for every frame $f_i \in V_{short}$, $R < i \leq W$, we compute the optical flow between frame $f_{i-R}$ and $f_i$. We then use the obtained optical flow for every pair of frames to calculate the HOOF feature. However, since we make use of ROI extraction, we do not compute the HOOF feature over the entire frame, but we compute it for each extracted ROI separately.

\begin{figure}
    \centering
        \includegraphics[width=0.4\textwidth]{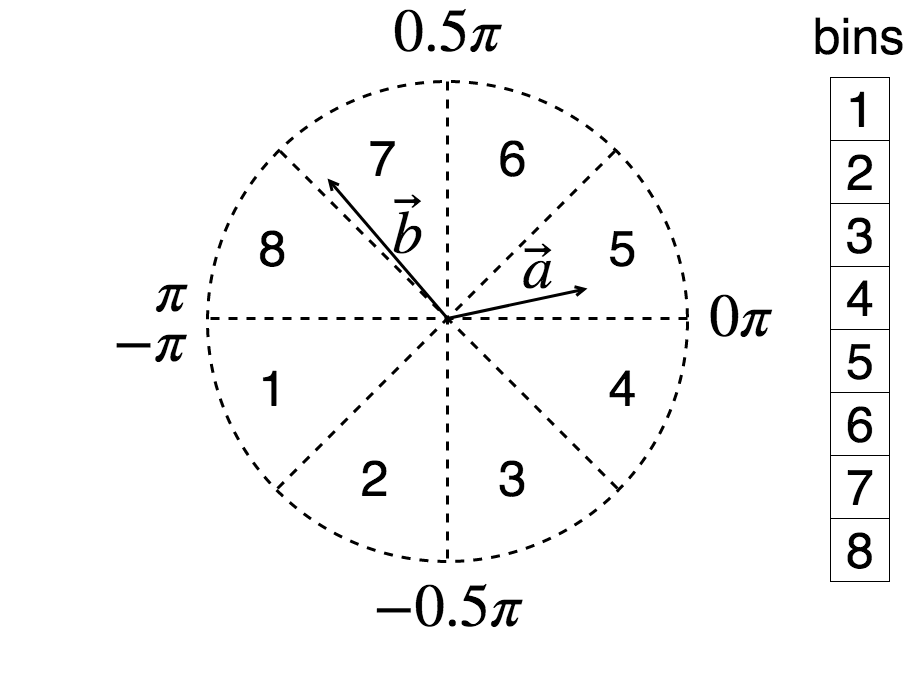}
    \caption{Histogram of Oriented Optical Flow feature with 8 bins. Flow vectors $\vec{a}$ and $\vec{b}$ are binned in bin 5 and 7 respectively, based on their angle.}
    \label{fig:HOOF}
\end{figure}

The Histogram of Oriented Optical Flow is calculated by binning flow vectors based on the orientation, and weighted by the magnitude of the vector. An example of how HOOF bins vectors is shown in Fig. \ref{fig:HOOF}. The original HOOF feature then sums the magnitudes of vectors in every bin and subsequently creates a normalized histogram from the orientation bins. Happy and Routray \cite{Happy2017} suggest a fuzzification method for HOOF, based on the motivation that vectors close to the border of a bin contribute nothing to the neighboring bin, meaning small differences can occasionally have a large impact on the results. Happy and Routray, therefore, suggest letting vectors contribute to multiple bins based on a membership function. This way, a vector like $\vec{b}$ in Fig. \ref{fig:HOOF} would contribute approximately equally to both bins 7 and 8. We take a similar approach, where we use soft binning implemented in the histogram function \cite{pdollar} of Piotr Dollar's Matlab Toolbox \cite{PMT}, which makes use of linear interpolation to let a vector contribute to the 2 closest orientation bins.

\subsection{Micro-expression spotting}
In this stage of the proposed method, we aim to determine whether a short interval of size $W$ contains a relevant micro-movement. Experimentation with handcrafted features demonstrates the difficulty of finding thresholds that allow micro-expressions to be reasonably distinguished from irrelevant facial movements. Another issue with thresholds is that they do not necessarily get better as we get more data. Therefore, we propose to use a recurrent neural network (RNN) consisting of long short-term memory (LSTM) units that learns to differentiate between relevant and irrelevant micro-movements. We choose to use an LSTM because it helps prevent the exploding and vanishing gradient problems associated with RNNs.

\begin{figure}
    \centering
        \includegraphics[width=0.4\textwidth]{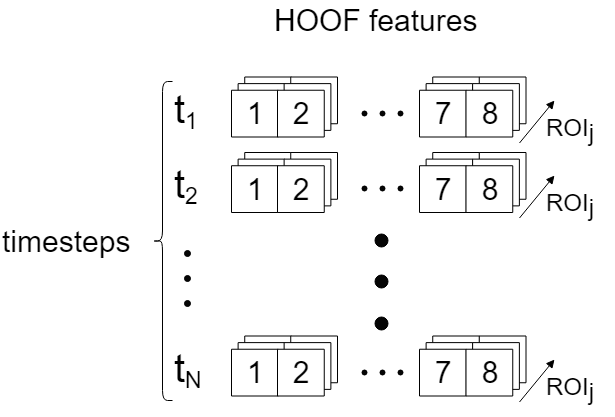}
    \caption{Example input sequence for the LSTM, consisting of $N=W/R$ timesteps, each containing a HOOF vector for every $ROI_j$, where $j \leq 3$, the number of ROIs we extracted.}
    \label{fig:input}
\end{figure}

The task for the LSTM is to classify whether a given sequence contains relevant micro-movements or not. The input sequences for the LSTM contain 25 timesteps, where timestep $t_i$ is based on the optical flow computation between frames $f_i$ and $f_{i+R}$. Therefore, each timestep has a HOOF feature for each of the extracted ROIs, as shown in Fig. \ref{fig:input}.

To train a neural network, we also need to pass the target values for every sample. The ground truth data of the SAMM dataset contains the onset, apex, and offset frame numbers. Whereas the input samples for the LSTM correspond to the short intervals of fixed size $W$. Therefore, we need to consider when the target value is true for a given interval of $W$ frames. We consider an interval $I_j$ to contain a micro-expression, i.e., the target value is true, if there exists an interval $S_{ME}$ in the ground truth data such that the condition in (\ref{eq:true}) holds.

The LSTM network consists of 2 LSTM layers, each with 12 dimensions. We chose to use 12 dimensions after trying other suitable amounts based on multiple rules-of-thumb methods \cite{Heaton2008}. The prediction for the binary classification is calculated using the softmax function. We use the ADAM optimizer in Keras, with a learning rate of $10^{-3}$ and a decay of $0$. We run the network for 50 epochs, which is sufficient for it to learn the representations without overfitting.

\subsection{Post-processing}
\label{sec:post-processing}
The LSTM network provides a list of predictions for the sequences that were generated based on the sliding window method described in Section \ref{sec:preprocessing}. However, the sliding window method is accompanied by an overlap $W_{overlap}$, and thus detections will overlap as well. To solve the issue of overlapping detections, we post-process the output of the network and suppress neighboring detections. We use a method similar to a greedy non-maximum suppression approach in object detection methods, where we take the interval with the maximum confidence among consecutive detections, and suppress the direct (overlapping) neighbors.

 
\section{RESULTS}
\label{sec:results}
\begin{figure}[thpb]
  \centering
  \includegraphics[width=0.45\textwidth]{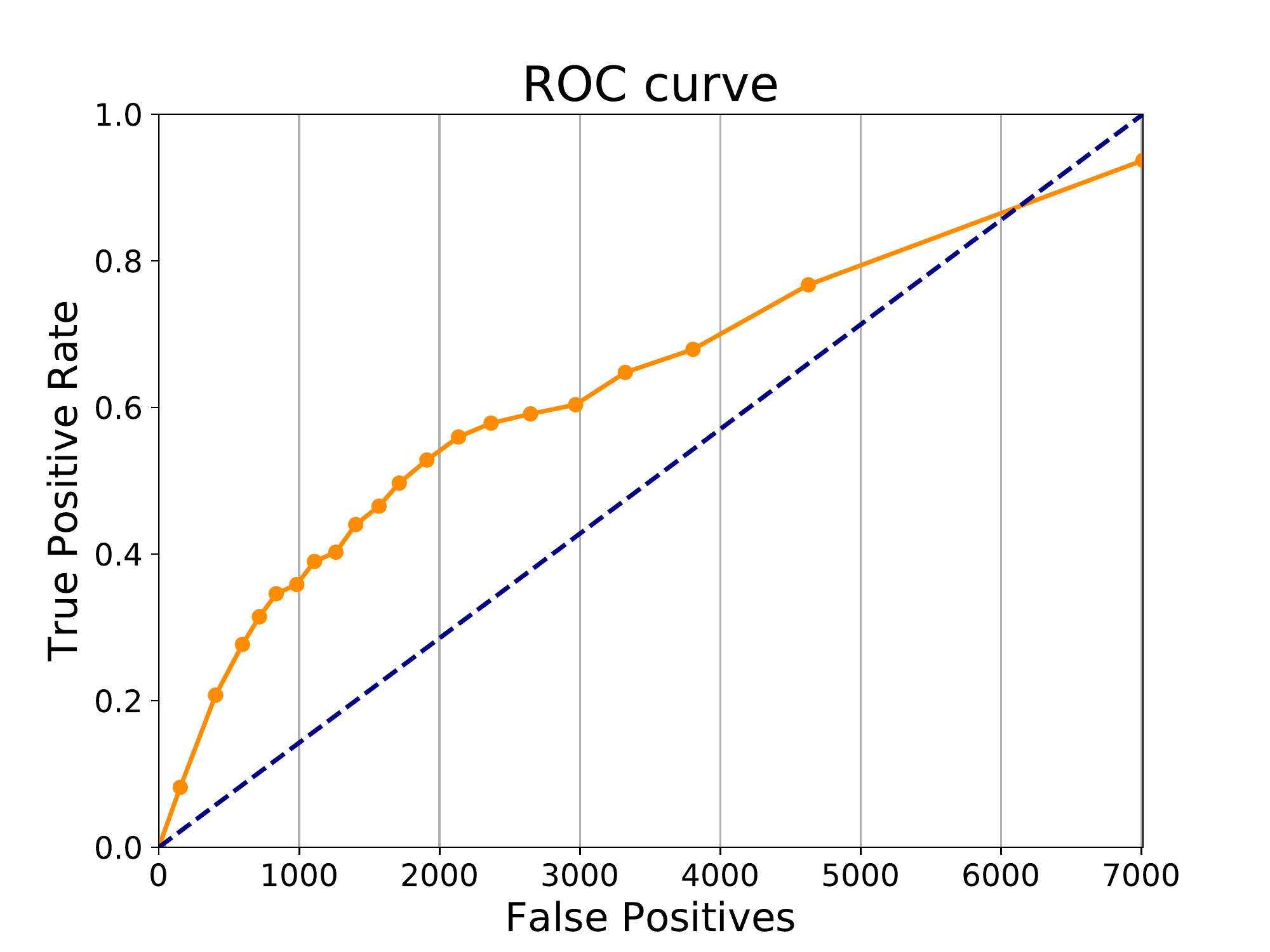}    
  \caption{The ROC curve for the proposed method on the SAMM dataset.}
  \label{fig:ROC}
\end{figure}

The proposed method is evaluated using the SAMM dataset, which contains 79 videos with 159 micro-movements in total. Since the method employs Leave-One-Subject-Out cross-validation, we ensure there is no subject bias in the computed results. A spotted interval $I_j$ is considered to be a true positive (TP) if there exists a micro-expression interval $S_{ME}$ such that the condition in (\ref{eq:true}) holds. Otherwise, if no such interval $S_{ME}$ exists, the spotted interval $I_j$ is considered a false positive (FP). Additionally, the number of false negatives (FN) is obtained by taking the total number of micro-expressions in the dataset, subtracted by the number of true positives. Our definition of a TP differs from the proposed definition of a TP in Li et al. \cite{Li2018a}, in which the interval needs to be in proportion to the duration of the micro-expression. The definition of a TP described in Li et al. would not work well for our method, because it currently focuses on spotting intervals of size $|W|$, which can be passed to annotators, rather than spotting the exact onset and offset of a micro-expression. 
Several metrics that are computed on the obtained results are displayed in Table \ref{tab:metrics}. Additionally, we provide the ROC curve for the proposed method in Figure \ref{fig:ROC}.

\begin{table}
\caption{Detection results for spotting micro-expressions intervals of 0.5s on the SAMM dataset.}
\label{tab:metrics}
\begin{center}
\begin{tabular}{|c|c|}
\hline
Recall & 0.4654\\
\hline
F1-score & 0.0821\\
\hline
Precision & 0.0450\\
\hline
TP & 74\\
\hline
FP & 1569\\
\hline
FN & 85\\
\hline
\end{tabular}
\end{center}
\end{table}

\section{DISCUSSION}
The proposed method performs quite well, albeit with a slightly different definition of a true positive. On all computed metrics it outperforms the baseline presented by Li et al. \cite{Li2018a}. To accurately compare to other state-of-the-art methods, we would need to use the exact same definitions. In our case case, to accomplish that, we would need to spot the onset and offset frames more precisely, which could be done in the post-processing phase. If we decrease the stride of the sliding window, which is currently quite large, we could localize where a micro-expression starts and ends based on the prediction scores of the overlapping intervals. Another method could be to use feature difference analysis on the short intervals of size $W$, to try and identify more precisely the onset and offset of the micro-expressions.

Furthermore, the network might have dataset bias, which would cause it to perform poorly on unseen samples. Hence, to get a more realistic idea about the performance of the method on unseen data, Composite Database Evaluation and Holdout-Database Evaluation could be performed \cite{Yap2018}.

When analyzing some of the results manually, one limitation of the method becomes clear. The issue occurs whenever a subtle head movement coincides with the occurrence of a micro-expression. The global head movement then influences the direction of the local movement of the micro-expression and thus changes the feature representation of the micro-expression, which in turn makes it harder for the neural network to recognize. The way we currently extract features, the information about the global head movement is largely lost, so there is no way to take it into account.

\section{CONCLUSIONS AND FUTURE WORKS}\label{sec:conclusion}
To conclude, this paper introduces a novel method that makes use of a long short-term memory network for spotting micro-expressions in long videos. The main contribution is that the method can support the annotation process for newly acquired data by reducing the length of footage to be reviewed while maintaining most of the micro-expressions. Indeed, the number of false positives indicate that an annotator would need to review only around a quarter of the length of the original footage, while still retaining about half of the micro-movements. Additionally, the neural network will improve as it gets more data from the annotation process. As a result, we hope that this method will eventually be able to help ease the burden of annotation. Especially because annotation is currently time-consuming because of the sparseness of the spontaneous data.

In the future, we hope to do a more elaborate analysis of the proposed method. Additionally, we would like to explore alternative approaches to various stages of the method like preprocessing, feature extraction and post-processing.


\bibliographystyle{IEEEtran}
\bibliography{references}

\end{document}